\documentclass[conference]{IEEEtran}
\usepackage{float}
\IEEEoverridecommandlockouts

\usepackage{cite}

\usepackage{amsmath,amssymb,amsfonts}
\usepackage{algorithmic}
\usepackage{graphicx}
\usepackage{textcomp}
\usepackage{xcolor}
\usepackage{float}
\usepackage{hyperref}
\def\BibTeX{{\rm B\kern-.05em{\sc i\kern-.025em b}\kern-.08em
    T\kern-.1667em\lower.7ex\hbox{E}\kern-.125emX}}
\begin{document}

\title{Extracting and Analyzing Rail Crossing Behavior Signatures from Videos using Tensor Methods

}

\author{
  \IEEEauthorblockN{
    Dawon Ahn\textsuperscript{1}\thanks{Dawon Ahn and Het Patel contributed equally to this work.},
    Het Patel\textsuperscript{1},
    Aemal Khattak\textsuperscript{2},
    Jia Chen\textsuperscript{3},
    Evangelos E. Papalexakis\textsuperscript{1}
  }
  \vspace{0.3em}
  \IEEEauthorblockA{
    \textsuperscript{1}Department of Computer Science and Engineering, UC Riverside, Riverside, USA\\
    \textsuperscript{2}Department of Civil and Environmental Engineering, University of Nebraska-Lincoln, Lincoln, USA\\
    \textsuperscript{3}Department of Electrical and Computer Engineering, UC Riverside, Riverside, USA\\
    \vspace{0.2em}
    \{dahn017, hpate061, epapalex\}@ucr.edu, khattak@unl.edu, jiac@ucr.edu
  }
}

\maketitle

\begin{abstract}
Railway crossings present complex safety challenges where driver behavior varies by location, time, and conditions. Traditional approaches analyze crossings individually, limiting the ability to identify shared behavioral patterns across locations. We propose a multi-view tensor decomposition framework that captures behavioral similarities across three temporal phases: Approach (warning activation to gate lowering), Waiting (gates down to train passage), and Clearance (train passage to gate raising). We analyze railway crossing videos from multiple locations using TimeSformer embeddings to represent each phase. By constructing phase-specific similarity matrices and applying non-negative symmetric CP decomposition, we discover latent behavioral components with distinct temporal signatures. Our tensor analysis reveals that crossing location appears to be a stronger determinant of behavior patterns than time of day, and that approach-phase behavior provides particularly discriminative signatures. Visualization of the learned component space confirms location-based clustering, with certain crossings forming distinct behavioral clusters. This automated framework enables scalable pattern discovery across multiple crossings, providing a foundation for grouping locations by behavioral similarity to inform targeted safety interventions.

\end{abstract}


\section{Introduction}
Crashes at railway crossings are a national concern in the US, with the majority of such crashes resulting from motorists failing to yield to oncoming trains~\cite{fra2016analysis}. Railway crossings present complex safety challenges where driver behavior varies by location, time of day, and environmental conditions. Traditional approaches analyze crossings individually, but this requires extensive resources and may miss opportunities to apply successful interventions across crossings that experience similar unsafe behaviors. Automated video analysis can identify patterns across multiple crossings, enabling domain experts to group locations for targeted interventions. 

Previous work in railway crossing safety has focused on individual location analysis or aggregate statistics across regions~\cite{salim2018naturalistic,LIANG2017181,ZHANG2018276,fra2016analysis}. However, these approaches do not capture the multifaceted nature of crossing events, in which driver behavior evolves across distinct temporal phases (warning, waiting, clearance) and varies across locations and times of day.

We address this challenge by representing railway crossing videos in a multi-view tensor framework, where each ``view" corresponds to behavioral similarity between all pairs of crossing events during a specific phase (approach, waiting, clearance).

\section*{Contributions}
\begin{itemize}
\item \textbf{Multi-View Behavioral Framework:} 
We introduce a multi-view tensor framework for railway crossing analysis that explicitly models behavioral similarities across three distinct temporal phases (approach, waiting, clearance), capturing how driver behavior evolves through crossing events.
\item 
\textbf{Interpretable Component Discovery:} We demonstrate that symmetric CP decomposition on a phase-specific similarity tensor successfully discovers interpretable behavioral components with distinct temporal signatures, validated through multiple rank selection metrics (CORCONDIA~\cite{CORCONDIA}, reconstruction error, hold-out validation).
\item \textbf{Cross-Location Analysis:} Through analysis of 31 crossing videos from 4 locations, we demonstrate that (1) crossing location seems to be a stronger determinant of behavioral patterns than time-of-day, and (2) approach-phase behavior provides particularly discriminative signatures. These findings provide initial evidence for phase-specific intervention strategies and crossing grouping for safety planning.
\end{itemize}

\section{Related Work}

Prior research on railway crossing safety has largely taken two forms: individual-location behavioral studies and aggregate regional crash analyses, with limited work examining shared behavioral structure across multiple crossings or across temporal phases of a crossing event. 

For example, Salim et al.~\cite{salim2018naturalistic} analyzed naturalistic driving videos to characterize how drivers respond to warnings at individual crossings, but their analysis remains crossing-specific and does not compare behavioral signatures across locations or phases. Liang et al.~\cite{LIANG2017181} conducted an observational analysis of the violation rates at different times of day and gate operation phases at four crossings in France, finding temporal patterns in risky behavior, but analyzing each crossing independently. Recent work has explored automated approaches: Zhang et al.~\cite{ZHANG2018276} use computer vision to automatically detect near-miss events of trespassing from surveillance video at grade crossings, and Amin et al.~\cite{amin2024intelligentrailroadgradecrossing} propose an intelligent crossing system using YOLO-based object detection and UNet segmentation to detect approaching trains and monitor crossing clearance. Although these studies provide valuable information on driver behavior patterns and automated detection at individual locations, they do not identify shared behavioral patterns that could enable grouping crossings for intervention planning.

\section{Methodology}

\subsection{Overview}
\begin{figure*}[h]
    \centering
    \includegraphics[width=0.8\linewidth]{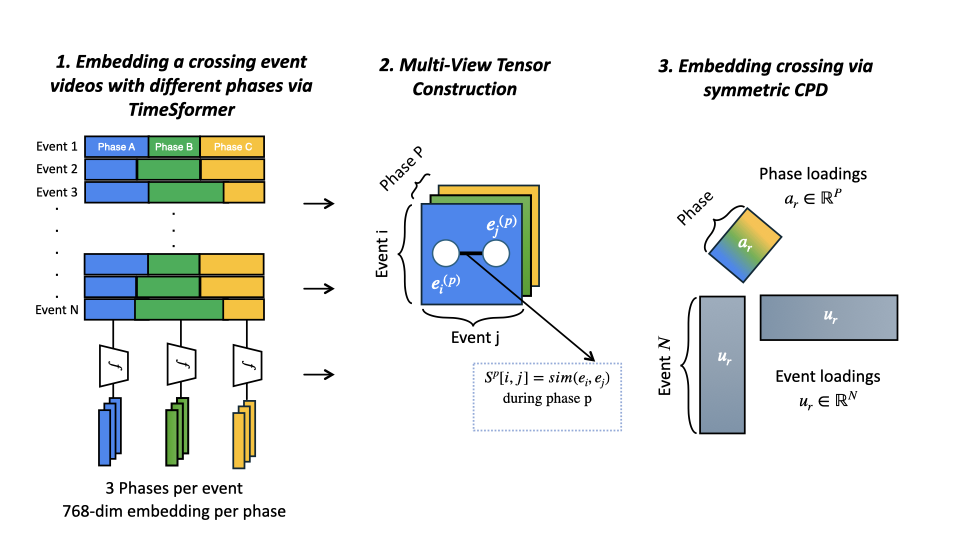}
    \caption{Overview of the proposed multi-view tensor decomposition pipeline. (1) Each crossing event video is segmented into three behavioral phases (Approach, Waiting, Clearance) and processed through TimeSformer to extract 768-dimensional embeddings per phase. (2) For each phase $p$, we construct a $31 \times 31$ similarity matrix by computing pairwise cosine similarity between video embeddings, then stack the three matrices to form a tensor $X \in \mathbb{R}^{31\times31\times3}$. (3) Symmetric non-negative CP decomposition factorizes the tensor into phase loadings $a_r \in \mathbb{R}^P$ and event loadings $u_r \in \mathbb{R}^N$, revealing latent behavioral components across videos and phases.}
    \label{fig:overview}
\end{figure*}

Figure \ref{fig:overview} provides an overview of our three-stage pipeline. First, we segment crossing videos into three behavioral phases and extract TimeSformer \cite{gberta_2021_ICML} embeddings for each phase (\ref{sec:Data}). Second, we construct a multi-view tensor by computing phase-specific similarity matrices and stacking them along the third dimension (\ref{sec:Embedding-Extraction}). Third, we apply symmetric CP decomposition \cite{Harshman1970FoundationsOT} to discover latent behavioral components, with rank selection validated through multiple diagnostics (\ref{sec: TensorConstruction} - \ref{sec: rank-selection}). The following subsections detail each stage. 

\subsection{Phase Annotation} \label{sec:Data}


Crossing videos are manually annotated into five behavioral phases:

\begin{enumerate}
    \item Pre-Event: Normal traffic before warning activation
    \item Phase A (Approach): Warning lights flash $\rightarrow$ Gate fully lowered.
    \item Phase B (Waiting): Gates down $\rightarrow$ Train clears the crossing.
    \item Phase C (Clearance): Train clears $\rightarrow$ Gates fully raised.
    \item Post-Event: Normal traffic after gates were raised.
\end{enumerate}

We focus our analysis on phases A, B, and C, as these represent the constrained event period during which the crossing mechanisms influence driver behavior. Videos are also labeled by time-of-day category to enable temporal pattern analysis. 


\subsection{Video Embedding Extraction} \label{sec:Embedding-Extraction}
For each phase in each video, we extracted 768-dimensional embeddings using TimeSformer (facebook/timesformer-base-finetuned-k400), a transformer-based video understanding model pre-trained on Kinetics-400.

\textbf{Multi-clip sampling strategy}: Rather than using a single frame or clip, we sampled multiple clips distributed evenly across each phase's duration to capture temporal dynamics:

\begin{itemize}
    \item Phase $< 20$ seconds: 1 clip
    \item Phase $20$--$60$ seconds: 3 clips
    \item Phase $> 60$ seconds: 5 clips
\end{itemize}

Each clip consisted of 8 frames sampled at a stride of 4 (approximately 1 second of video at 30fps). The final embedding for each phase was computed as the mean of all clip embeddings, capturing behavioral dynamics across the entire phase rather than a single moment. This resulted in a total of 93 phase embeddings: 31 videos × 3 phases (A, B, C).

\subsection{Multi-View Tensor Construction} \label{sec: TensorConstruction}
For each phase $p \in \{A, B, C\}$, we computed the pairwise cosine similarity between all video embeddings: 

$$
S^{(p)}_{ij} = \frac{e^{(p)}_i \cdot e^{(p)}_j}{\|e^{(p)}_i\| \|e^{(p)}_j\|}
$$

where $e^{(p)}_i$ is the embedding for video i during phase p.

This produced three 31x31 symmetric similarity matrices, which we stacked into a third-order tensor: 
$$
\mathcal{X} \in \mathbb{R}^{N \times N \times P}
$$
where N=31 videos and P=3 phases.

Each frontal slice $\mathcal{X}_{p,:,:}$ represents behavioral similarities during phase p. This multi-view representation naturally captures how behavioral patterns differ across phases while maintaining the video-level structure.

\subsection{Rank Selection} \label{sec: rank-selection}
Selecting the appropriate rank R for tensor decomposition is critical. Too low and important patterns are missed; too high, and the model overfits. We utilized three complementary metrics

\subsubsection{CORCONDIA (Core Consistency Diagnostic)}
CORCONDIA \cite{CORCONDIA} measures how well the data conform to the CP/PARAFAC structure: 

$$
\text{CORCONDIA} = 100 \times \left(1 - \frac{\|\mathcal{G} - \mathcal{I}\|^2_F}{R}\right)
$$

where $\mathcal{G}$ is the Tucker \cite{Tucker_1966} core obtained by projecting the tensor onto the CP factor spaces, $\mathcal{I}$ is the superdiagonal identity tensor, and R is the rank. Values above 80\% indicate good CP fit. CORCONDIA is valid only when R $\leq$ min(I, J, K).

\begin{figure}[H]
    \centering
    \includegraphics[width=1\linewidth]{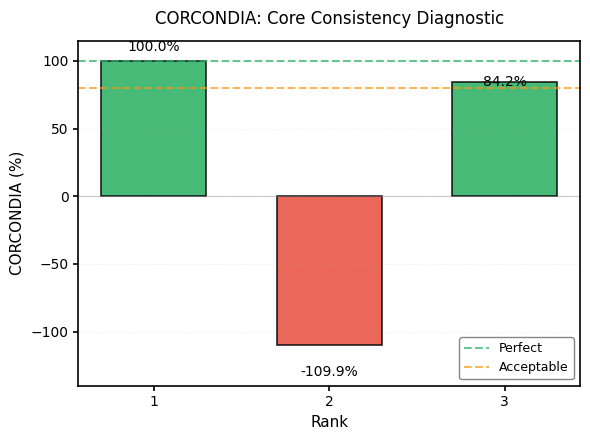}
    \caption{CORCONDIA diagnostic for rank selection. CORCONDIA measures core consistency, 
where values $\ge$ 80\% (orange line) indicate acceptable CP structure and $\approx$ 100\% (green line) 
indicates a perfect fit. Rank 2 shows severe structural issues (-109.9\%), while ranks 1 and 
3 demonstrate a valid CP structure. Only valid for R $\leq$ min(I,J,K) = 3.}
    \label{fig:CORCONDIA}
\end{figure}

\subsubsection{Reconstruction Error}
We measured the sum of squared differences between the original tensor and its CP reconstruction across ranks 1 through 10. The error decreased consistently, with diminishing returns observed after rank 3-4.


\begin{figure}[ht]
    \centering
    \includegraphics[width=1.0\linewidth]{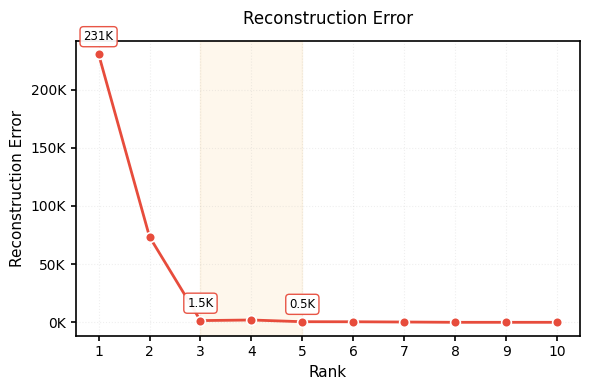}
    \caption{Reconstruction error (sum of squared errors) across ranks 1-10. The curve shows 
a clear elbow between ranks 3-5 (shaded region), indicating diminishing returns beyond 
rank 3. Values shown in thousands (K)}
    \label{fig:Reconstruction}
\end{figure}

\subsubsection{Holdout Validation}
We randomly mask 10\% of the tensor entries and measure the prediction error on held-out entries. We used TensorLy's \cite{kossaifi2019tensorly} mask parameter during decomposition to prevent information leakage (treating masked entries as unobserved rather than zero). We performed three trials with five random restarts each to ensure stability. 

Holdout RMSE: 

\begin{itemize}
    \item Rank 1: 80.97 ± 10.26
    \item Rank 2: 63.54 ± 19.96
    \item Rank 3: 37.01 ± 31.64
    \item Rank 4: 41.90 ± 32.55
\end{itemize}

\begin{figure}[h]
    \centering
    \includegraphics[width=1.0\linewidth]{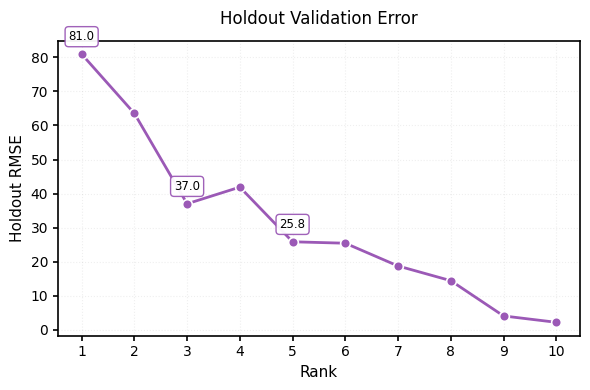}
    \caption{Holdout validation error (RMSE) with 10\% masked entries across ranks 1-10. 
Lower values indicate better generalization. Error generally decreases with rank, with substantial improvements up to rank 5. Proper masking prevents information leakage during training (5 random restarts per trial, averaged over 3 trials).}
    \label{fig:holdout}
\end{figure}

Based on these metrics, ranks 3, 4, and 5 all represent reasonable choices, with CORCONDIA supporting rank 3, reconstruction error showing diminishing returns in the rank 3-5 range, and holdout validation showing comparable performance across these ranks. We selected rank 4, as empirical comparison of the decompositions showed that rank 3 did not surface meaningful within-location variability, while rank 5 introduced components with substantial overlap that were difficult to distinguish. Rank 4 provided the best balance of interpretability and expressiveness.  

\subsection{Non-Negative Symmetric CP Decomposition}

We applied non-negative symmetric CP decomposition to factor the tensor \cite{Harshman1970FoundationsOT}:
$$
\mathcal{X} \approx \sum_{r=1}^{R} \lambda_r \, \mathbf{a}_r \circ \mathbf{u}_r \circ \mathbf{u}_r
$$
\begin{itemize}
    \item $\lambda_r \in \mathbb{R}^+$ is the non-negative scalar weight for component $r$
    \item $\mathbf{a}_r \in \mathbb{R}^{3}_+$ are non-negative phase loadings for component $r$ (how much each phase contributes)
    \item $\mathbf{u}_r \in \mathbb{R}^{31}_+$ are non-negative video loadings for component $r$ (how much each video contributes)
    \item $\circ$ denotes outer product
\end{itemize}

The symmetric structure ($\mathbf{u}_r$ appears twice) reflects the symmetric nature of similarity matrices. The non-negativity constraint ensures all factor loadings are non-negative, making each video interpretable as a non-negative mixture of behavioral components—a representation more intuitive to interpret than standard CP decomposition's bipolar structure.

We optimized using alternating least squares with 2000 iterations, performing 5 random restarts, and selecting the solution with the lowest reconstruction error.

\section{Results}

\begin{figure}
    \centering
    \includegraphics[width=0.78\linewidth]{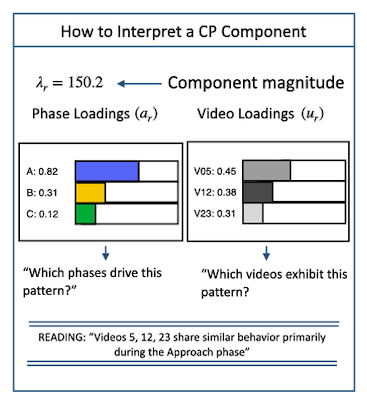}
    \caption{Reading a CP decomposition component. Component magnitude ($\lambda_r$) indicates overall importance. Phase loadings ($a_r$) reveal which phases (Approach, Waiting, Clearance) define the behavioral pattern. Video loadings ($u_r$) identify which crossing events exhibit this pattern. The example component is Approach-dominant (0.82) and characterizes videos 5, 12, and 23.}
    \label{fig:interp}
\end{figure}

\subsection{Dataset and Distribution}

We analyzed 31 railway crossing videos collected in February 2024 from 4 locations in Lincoln, Nebraska.

\begin{itemize}
    \item \textbf{35\textsuperscript{th} Street and Cornhusker Highway:} 23 videos
    \item \textbf{NW 12\textsuperscript{th} Street and Cornhusker Highway:} 6 videos
    \item \textbf{27\textsuperscript{th} Street and Nebraska Pkwy:} 1 video
    \item \textbf{56\textsuperscript{th} Street and Old Cheney:} 1 video
\end{itemize}

The videos were labeled by time-of-day category: off-peak (7 PM–6 AM, 12 videos), morning rush (6 AM–10 AM, 9 videos), midday (10 AM–3 PM, 8 videos), and afternoon/evening (3 PM–7 PM, 2 videos).

\begin{figure}[h]
    \centering
    \includegraphics[width=1\linewidth]{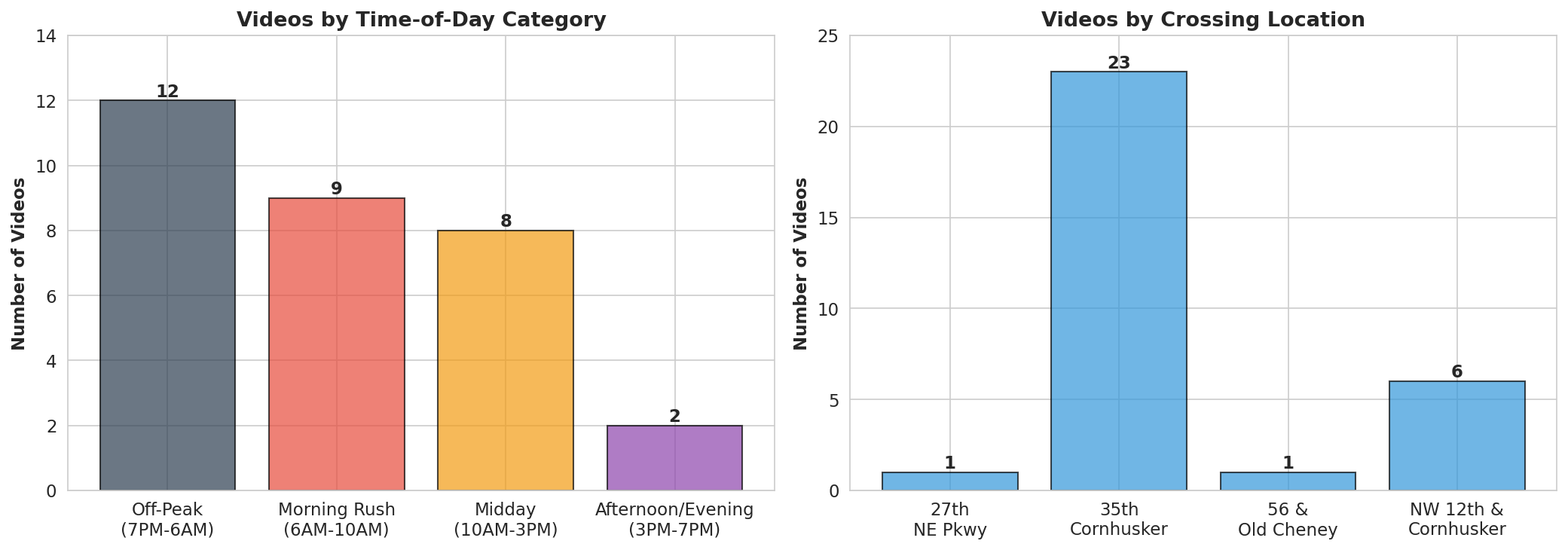}
    \caption{Distribution of 31 videos across crossing locations and time-of-day categories.}
    \label{fig:video_distribution}
\end{figure}

Figure \ref{fig:video_distribution} shows the distribution of 31 videos across crossing locations and time-of-day categories. The 35\textsuperscript{th} Street and Cornhusker Highway location accounts for the majority of videos (23), providing sufficient samples to examine within-location behavioral variability.

\subsection{Multi-View Similarity Tensor}

\begin{figure}[h]
\centering
\includegraphics[width=0.48\textwidth]{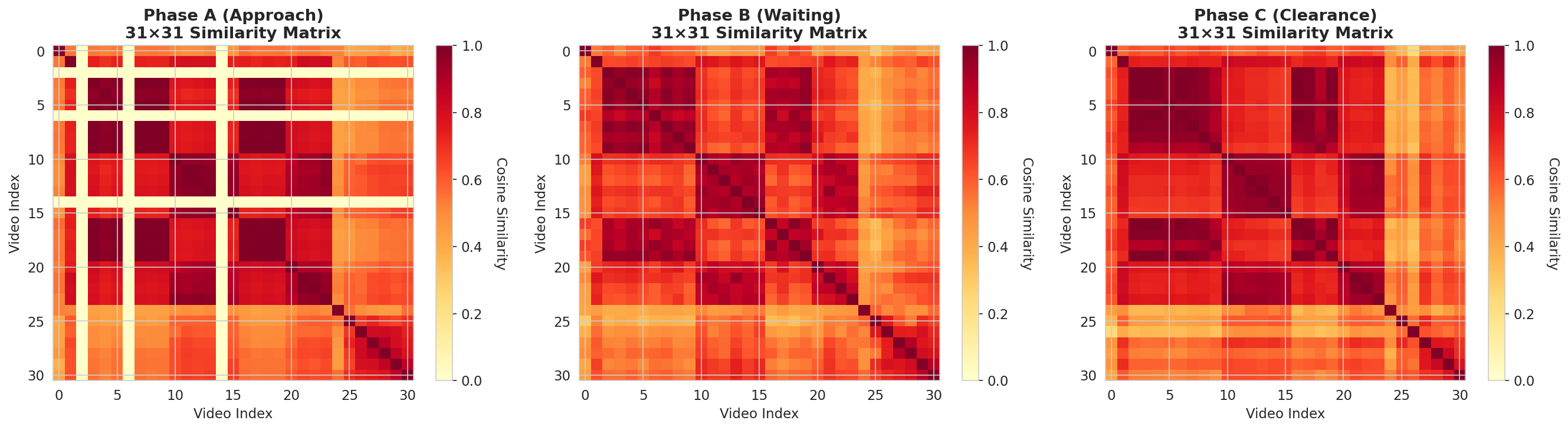}
\caption{Multi-view similarity tensor. Each $31\times31$ matrix shows pairwise cosine similarity between videos for one phase. Darker red indicates higher similarity.}
\label{fig:similarity_matrices}
\end{figure}

Figure \ref{fig:similarity_matrices} shows the three $31\times31$ similarity matrices for phases A (Approach), B (Waiting), and C (Clearance). Each matrix exhibits a distinct structure, with Phase A displaying more pronounced block patterns, indicating clearer behavioral groupings during the approach period.

\subsection{Non-Negative Symmetric CPD (Rank 4)}

\begin{figure}[h]
    \centering
    \includegraphics[width=1\linewidth]{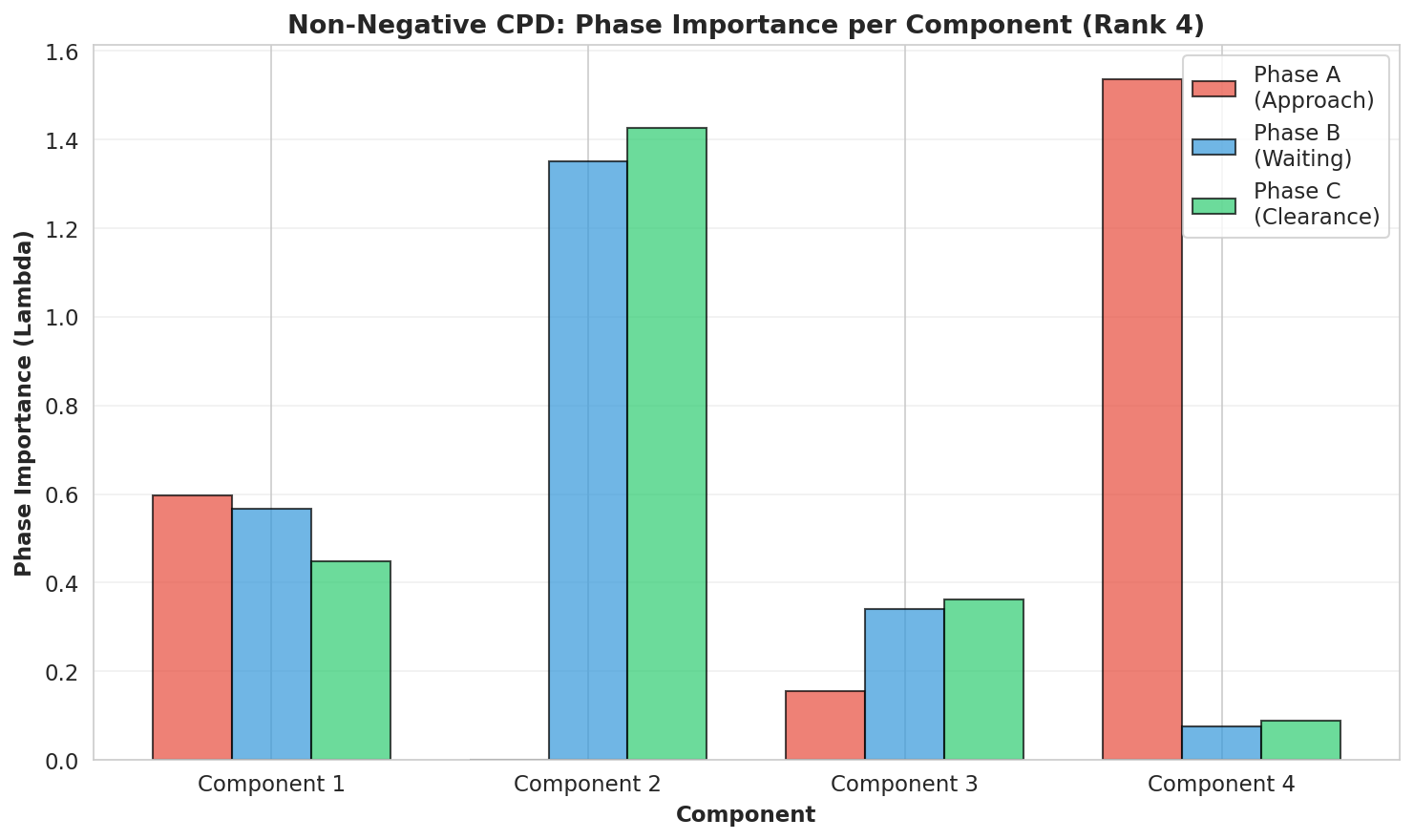}
    \caption{Phase loadings ($a_r$) for non-negative symmetric CPD at rank 4. Each grouped bar shows the contribution of the three phases (Approach, Waiting, Clearance) to one of the four latent components.}
    \label{fig:nonnegative_phase_importance}
\end{figure}

Phase loadings (Figure \ref{fig:nonnegative_phase_importance}) reveal different temporal signatures for each component. \textbf{Component 4} exhibits strong dominance of the approach-phase (1.52), indicating that the initial driver response to crossing warnings provides a particularly discriminative behavioral signature. \textbf{Component 2} emphasizes the waiting and clearance phases (1.34 and 1.43) with minimal approach contribution, capturing post-gate-lowering behavior. \textbf{Component 3} shows moderate engagement across all phases (0.16-0.36). \textbf{Component 1} displays balanced contributions across phases (0.45-0.60).

\begin{figure}[h]
    \centering
    \includegraphics[width=1\linewidth]{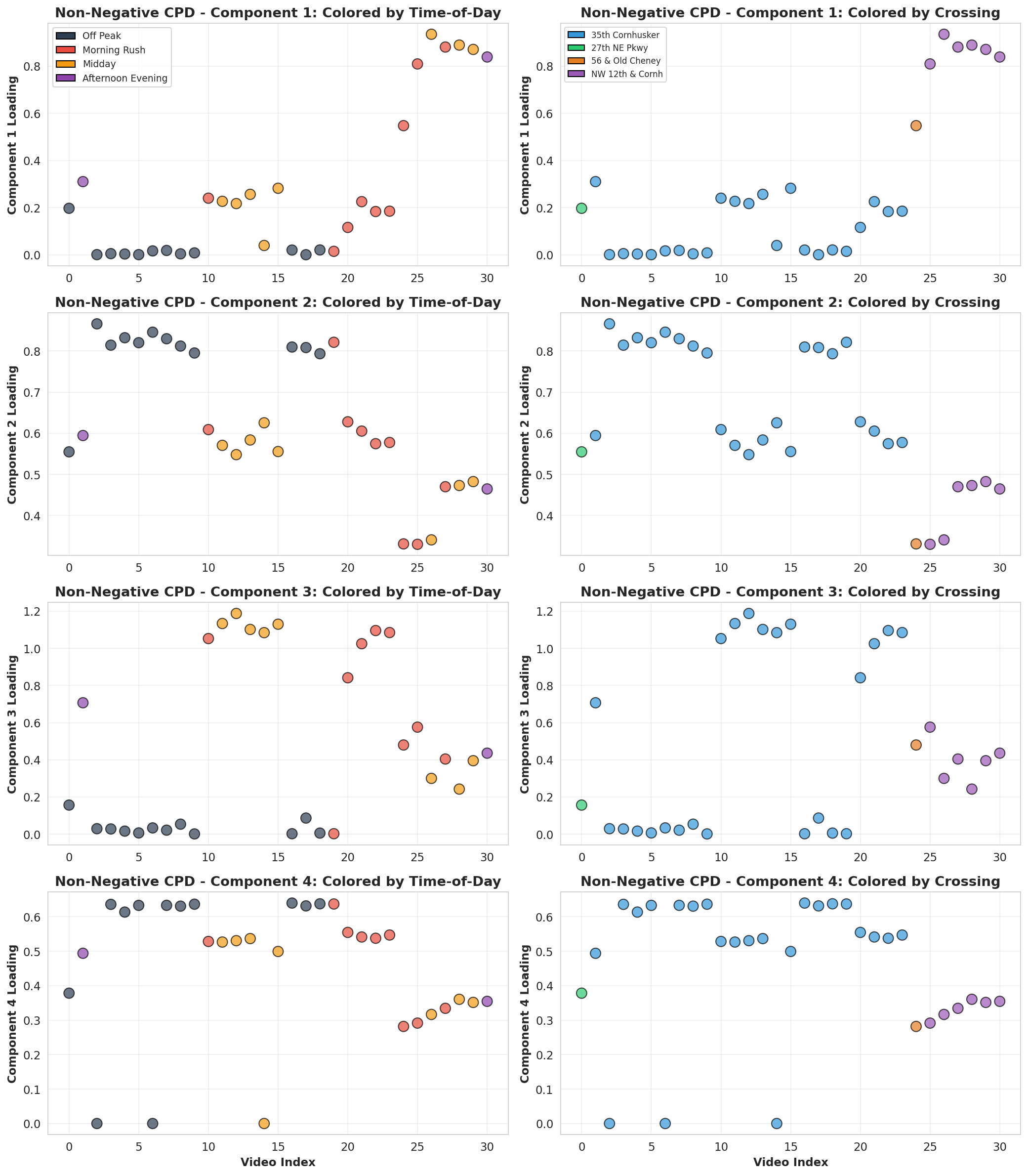}
    \caption{Non-negative CPD (rank 4) video loadings. Left: colored by time-of-day. Right: colored by crossing location.}
    \label{fig:nonnegative_cpd}
\end{figure}

Figure \ref{fig:nonnegative_cpd} shows the video loadings across all 4 components, colored by time-of-day (left) and crossing location (right). The location-based patterns (right panels) show a much clearer structure than the temporal patterns. \textbf{Component 1} exhibits the strongest location separation: NW 12\textsuperscript{th} Street videos (purple) cluster at high loadings (0.8-0.9), while 35\textsuperscript{th} Street videos (blue) remain near zero (0.0-0.3). Given Component 1's balanced phase loadings, this suggests that NW 12\textsuperscript{th} Street exhibits distinctive characteristics throughout the entire crossing event. \textbf{Component 2} shows 35\textsuperscript{th} Street videos with higher engagement (0.55-0.9) compared to NW 12\textsuperscript{th} Street (0.35-0.5). \textbf{Component 3} reveals substantial heterogeneity within the location: 35\textsuperscript{th} Street videos span a wide range (0.0-1.2), indicating that not all events at the same crossing exhibit identical behavioral patterns. \textbf{Component 4} shows a more distributed engagement between locations (0.0-0.65).

The time-of-day patterns (left panels) show substantial overlap across components. Component 3 exhibits some temporal structure, with off-peak videos clustering at low loadings (0.0-0.2), while other time periods show higher, more variable engagement (0.4-1.15). However, Components 2 and 4 show no clear temporal separation, with all time categories intermixed throughout the loading range. This pattern suggests that the location of the crossing is the dominant factor in the structuring of behavioral signatures, while temporal factors contribute to secondary variation. The substantial variability within-location observed in Component 3 suggests that factors beyond crossing location and time-of-day—potentially including specific traffic conditions or situational variables - also influence behavior patterns.

\subsection{t-SNE Visualization of Component Space}

To visualize the 4-dimensional component space learned by non-negative CPD, we applied t-SNE dimensionality reduction (perplexity=5, learning rate=200, 1000 iterations) to project the 31 video embeddings into 2D.

\begin{figure}[h]
    \centering
    \includegraphics[width=0.48\textwidth]{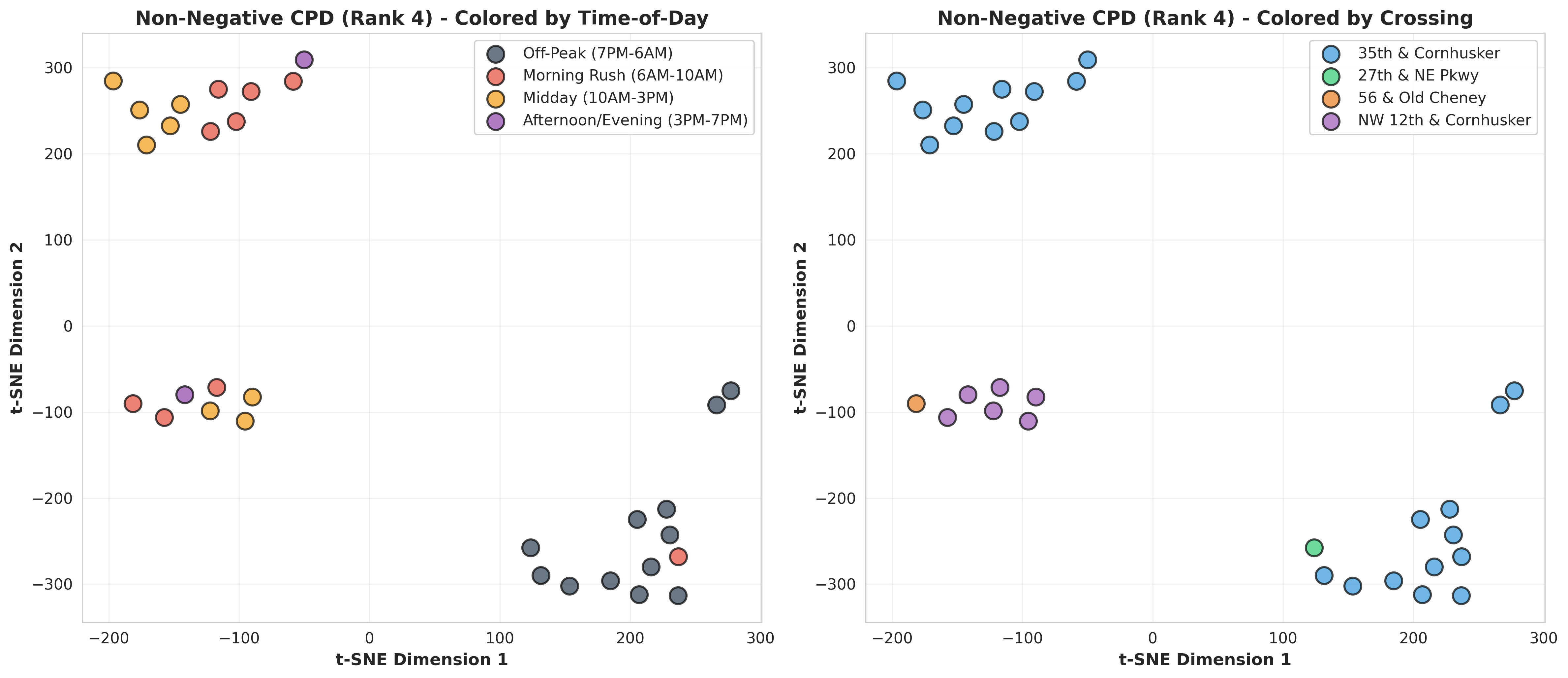}
    \caption{t-SNE projection of non-negative CPD component space. Left: colored by time-of-day. Right: colored by crossing location.}
    \label{fig:tsne_ncpd}
\end{figure}

Figure \ref{fig:tsne_ncpd} shows the t-SNE projection for non-negative CPD. The right panel (colored by crossing) reveals location-based clustering patterns. The 35\textsuperscript{th} Street and Cornhusker Highway videos (blue) separate into distinct subgroups: a tight upper cluster (y $\approx$ 200-300), a lower-right cluster (y $\approx$ -200 to -300), and isolated points. NW 12\textsuperscript{th} Street videos (purple) form a cluster in the middle-left region (x $\approx$ -150 to -50, y $\approx$ -100), partially overlapping the single videos from 27\textsuperscript{th} Street (green) and 56\textsuperscript{th} Street (orange).

The left panel (colored by time-of-day) shows substantial overlap across all temporal categories, with off-peak (dark gray), morning rush (red), midday (orange), and afternoon/evening (purple) videos intermixing throughout the component space. The lower-right cluster shows some concentration of off-peak videos, but the overall pattern suggests that crossing location is the dominant factor structuring the learned behavioral space, with temporal factors playing a secondary role. This location-based separation enables domain experts to group crossings by behavioral similarity for targeted safety interventions.

\section{Summary of Findings}
Our multi-view tensor decomposition analysis of 31 railway crossing videos reveals several key findings:

\textbf{Location Effects vs. Time-of-Day:} Within our samples, the t-SNE visualization and component loadings show clear location-based clustering, while time-of-day categories exhibit substantial overlap. NW 12\textsuperscript{th} Street forms a distinct behavioral cluster (Component 1 dominance), while 35\textsuperscript{th} Street videos distribute across Components 2-4, indicating within-location heterogeneity.

\textbf{Approach-Phase Discriminability:} Component 4's strong approach-phase dominance (loading: 1.52) demonstrates that initial driver response to crossing warnings provides a particularly discriminative behavioral signature, suggesting that approach-zone interventions may be especially impactful.

\textbf{Phase-Specific Patterns:} The decomposition successfully separates behavioral signatures by temporal phase, with Component 2 emphasizing waiting/clearance phases (loadings: 1.34/1.43) and Components 1 and 3 showing more balanced or moderate contributions.

\textbf{Within-Location Variability:} Component 3 reveals substantial heterogeneity within the 35\textsuperscript{th} Street location (loadings: 0.0-1.2), indicating that factors beyond the location of the crossing, such as specific traffic conditions or situational variables, influence behavioral patterns.

\section{Discussion and Limitations}

Our multi-view tensor decomposition framework enables automated behavioral pattern discovery across railway crossings. The observation that the location of the crossing appears to have stronger effects than time-of-day in our sample suggests that infrastructure modifications may be more impactful than temporal interventions. This enables practical applications: crossings with similar component profiles can be grouped for shared interventions (e.g. enhanced early warning systems for approach-phase dominant crossings), while crossings forming distinct clusters (e.g. NW 12\textsuperscript{th} Street) can be flagged for location-specific expert review. As railway agencies implement video monitoring systems, this automated approach complements traditional manual assessments.

Although the method successfully identifies location-based behavioral clusters, translating these patterns into specific intervention recommendations requires addressing several considerations:

\textbf{Lack of Crossing Characteristic Data:} While our method successfully identifies location-based behavioral clustering, we lack data on crossing characteristics (geometry, signage, traffic volumes, speed limits) that would explain why these behavioral differences occur. The tensor decomposition demonstrates that NW 12\textsuperscript{th} Street exhibits a distinct behavioral signature, but without infrastructure metadata, we cannot determine which physical features drive this distinction. Future work integrating crossing characteristic data could link discovered behavioral patterns to specific design or operational factors, enabling more targeted intervention recommendations.

\textbf{General-Purpose Video Model:} We use TimeSformer embeddings pre-trained on Kinetics-400, a general action recognition dataset. While this transfer learning approach successfully captures behavioral patterns sufficient for location-based clustering, fine-tuning on annotated railway crossing data could potentially improve sensitivity to domain-specific behaviors such as gate violations or near-miss events. The current embeddings effectively distinguish crossing locations, demonstrating that general video models provide useful representations even without domain-specific training.

\textbf{Limited Location Samples:} Our analysis spans 4 crossing locations with unbalanced sampling. Although location-based clustering is clear within these samples, validation is needed across a broader set of crossings to confirm the generalization of location-dominant behavior patterns. 

\section*{Acknowledgment}
Research was supported by the National Science Foundation under grant no. 2431569 and by the University Transportation Center for Railway Safety (UTCRS) at UTRGV through the USDOT UTC Program under Grant No. 69A3552348340. Research was also supported in part by the National Science Foundation under CAREER grant no. IIS 2046086, and CREST Center for Multidisciplinary Research Excellence in CyberPhysical Infrastructure Systems (MECIS) grant no. 2112650.

\bibliographystyle{plain}
\bibliography{references}

\end{document}